# Understanding the hand-gestures using Convolutional Neural Networks and Generative Adversial Networks

**Arpita Vats**
Department of Computer Science
Boston University
arpita8@bu.edu

## Abstract

In this paper, it is introduced a hand gesture recognition system to recognize the characters in the real time. The system consists of three modules: real time hand tracking, training gesture and gesture recognition using Convolutional Neural Networks. Camshift algorithm and hand blobs analysis for hand tracking are being used to obtain motion descriptors and hand region. It is fairy robust to background cluster and uses skin color for hand gesture tracking and recognition. Furthermore, the techniques have been proposed to improve the performance of the recognition and the accuracy using the approaches like selection of the training images and the adaptive threshold gesture to remove non-gesture pattern that helps to qualify an input pattern as a gesture. In the experiments, it has been tested to the vocabulary of 36 gestures including the alphabets and digits, and results effectiveness of the approach.

## 1 Introduction

Gesture recognition is an area of active current research in computer vision. It brings visions of more accessible computer system. In this paper, it focuses on the problem of hand gesture recognition using a real time tracking method with Camshift algorithm. It has been considered single- handed gestures, which are sequences of distinct hand shapes and hand region. A Gesture is defined as a motion of the hand to communicate with a computer. Many approaches to gesture recognition have been developed. A large variety of techniques have been used for modeling the hand. An approach based on the 2-D locations of fingertips and palms was used by Davis and Shah. Bobick and Wilson have developed dynamic gestures have been handled using framework. A state-based technique for recognition of gestures in which the define the feature as a sequence of states in a measurement or configuration space. Use of inductive learning for hand gesture recognition has also been explored. Yoon et al. have proposed a recognition scheme using combined features of location, angle and velocity. Lockton et al. propose a real time gesture recognition system, which can recognize 46 ASL, letter spelling alphabet and digits. The gestures that are recognized are static gesture of which the hand gestures do not move.
Several system use Hidden Markov models for gesture recognition. This research is focused on the application of the CNNs method to hand gesture recognition and GANs to understand the structure of the movements to predict the forthcoming gestures. The basic idea lies in the real-time generation of gesture model for hand gesture recognition in the content analysis of video sequence from CCD camera. Since hand images are two- dimensional and only a portion of the frames is taken to recognize the character, simple ConvNets are enough for recognition, and it will be helpful and offer a great potential for analyzing and recognizing gesture patterns. In addition, this gesture recognition system uses both the temporal and characteristics of the gesture for recognition. Unlike most other schemes, the system is robust to background clutter, does not use special glove to be worn and yet runs in real time. Although use to the knowledge for recognition is not new but this approach first time is introduced to the task of hand gesture recognition. Use of both hand regions, features of location, angle, and velocity and motion pattern are also novel feature in this work. ??The organization of the rest of the paper is as follows. Section 2 describes overview of the gesture recognition scheme. In Section 3, we discuss the tracking framework. We discuss hand gesture training and gesture recognition method in Section 4. The next section presents results of experiments and finally. The summarize the contribution of this work and identify areas for further work in the conclusion section.

## 2 Gesture Recognition overview

Several authors have emphasized the importance of using many diverse training examples for CNNs. They have proposed data augmentation strategies to prevent CNNs from overfitting when training with datasets containig limited diversity. Krizhevsky et al. employed translation, horizontal flipping and RGB jittering of the training and testing images for classifying them into 1000 categories. Simonyan and Zisserman employed similar spatial augmentation on each video frame to train CNNs for video-based human activity recognition. However, these data augmentation methods were limited to spatial variations only. To add variations to video sequences containing dynamic motion, Pigou et al. temporally translated the video frames in addition to applying spatial transformations. To reduce potential overfitting and improve generalization of the gesture classifier, it's proposed an effective spatio-temporal data augmentation method to deform the input volumes of hand gestures. The augmentation method also incorporates existing spatial augmentation techniques. This work bears similarities to the multi-sensor approach of Molchanov et al., but differs in the the use of two separate sub-networks and data augmentation. It is demonstrated that the system, with two sub-networks, that employs spatio-temporal data augmentation for training, outperforms both a single CNN and the baseline feature-based algorithm.

## 3 Implementation

We use a convolutional neural network classifier for dynamic hand gesture recognition. Sec. 3.1-3.4, describes the preprocessing steps needed for the model, the details of the classifier and the training of the two sub-networks. Finally, it is introduced a way to incorporate the GANs to predict the gesture form the initial movements.

### 3.1 Preprocessing

Each hand gesture sequence has a different time duration. To normalize the temporal lengths of the gestures, each gesture sequence is re-sampled to 32 frames using nearest neighbor interpolation by dropping or replacing frames. The original intensity and the depth of the images are also spatially down sampled by the factor of 2 to 28 x 28 pixels. The gradients are being computed from the intensity channel using the Sobel and Canny Operator of size 3 x 3 pixels. Gradients helped to improve robustness to the different illumination conditions. Each channel of the particular gesture's video sequence is normalized to be of zero mean and unit variance. This helps the classifier converge faster. The final inputs to the gesture classifier were 28 x 28 x 3

### 3.2 Segmentation

The main goal of the project is to make digits recognition processing real time colored-images and also using pre-stored videos or images. Therefore color recognition is the first and more important feature, and it will allow to perform all the techniques of the program. The first method consist on identifying the static background of an image, and isolate its foreground, allowing to replace the background with another image or video. On the first attempts this function was implemented using a "brute force" algorithm, it works iterating through all the pixels of each image, checking it to see if it fits inside the color range we have to detect, and check if it is part of the background or the foreground. The main problem founded in the tests of this algorithm was that it is very inefficient, and was impossible to keep the correct frame rate to made the calculations in real time. Therefore, in terms of efficiency, this algorithm is so slow, however it does a good color detection. Because of that, it was decided to compute the color detection in another way to try to improve the performance.
The next step to try to perform the efficiency was to use matrix operations to isolate the zones with the selected color. In this new algorithm, the absolute difference between the main image and a matrix filled completely with the selected color is calculated. This difference matrix will contain small values in the pixels with a color closer to the selected one, then all the pixels of the background will have a lower value than the threshold. This difference matrix is converted to gray-scale, and then compared with the threshold. A binary function is used to this calculation, setting the pixels with lower value than the threshold to black, and the rest to white.

$$dst(x, y) = \begin{cases} maxValue & \text{if } src(x, y) > threshold \\ 0 & \text{otherwise} \end{cases}$$

Then it's only needed to use this matrix as a mask in the next two operations to reach the final image. In the first step, a logic AND operation using the mask is applied to the main image with itself, and then the background pixels are removed. So now, as a second step, the image with the new background is added to the previous result using the same mask, and the pixels of the new background will added only to the black pixels where the old background was removed. The results of this method were almost perfect in the tests, replacing nearly all the pixels of the background if the threshold is correctly chosen. The efficiency of this algorithm also allows to perform the calculations in real time. This algorithm works comparing each frame of the image with a previous frame, trying to detect which zones stay the same between both frames, and assign those like the background. It does not work using colors to detect the background, instead of this it detects when the foreground changes using his movement. The main problem of this algorithm is that if the difference between two consecutive frames is not big enough, it does not detect the background good, so it only works in situations where the subject on the image moves. This algorithm is not so good in background detection, but could be useful to detect the movement in a scene when the background is static and something comes inside.

## 3.3 Tracking

The tracking algorithm is the second main functionality and it consists on detecting an object with a determinate color inside the scene, and track his position making possible to paint with it as a brush, or even use it as a pointer. A binary image which determines the shape of the object is used to know its exact location on the screen. Once this shape is isolated, is needed to calculate its center using the central moment (gravity center), a spot is drawn at the center of the object, and when this is performed along all the frames of a video, a path will be drawn, creating the sensation of painting. The algorithms used here are the camshift algorithm and the image differencing. CamShift is a tracking algorithm, which is based on a MeanShift Algorithm and it essentially does the function of the MeanShift in every single frame of the video which is executed in three steps: Back projection, MeanShift and Tracking. Back Projection uses the histogram of an image to show up the probabilities of colors may appear in each pixel. This function actually calculates the weight of each color in the whole picture using histogram, and changes the value of each pixel to the weight of its color in whole picture. MeanShift algorithm finds modes in a set of data samples representing an underlying probability density function (PDF) in $R^N$. It is a non-parametric clustering technique which does not require prior knowledge of the number of clusters, and does not constrain the shape of the clusters. Additionally, the detection and tracking is established using the moments of the consecutive frames. The moments of the scalar(greyscale) image with pixel intensities $I(x, y)$ are calculated by

$$M_{ij} = \sum_x \sum_y x^i y^i I(x,y)$$

And the central moments for the digital image $f(x,y)$ are defined as

$$\mu_{pq} = \sum_x \sum_y (x - \bar{x})^p (y - \bar{y})^q f(x,y)$$

where $\bar{x}$ and $\bar{y}$ are given by

$$\bar{x} = \frac{M_{10}}{M_{00}}, \bar{y} = \frac{M_{01}}{M_{00}} \text{ are the centroid}$$

### 3.4 Classifier

The convolutional neural network classifier consisted of two sub-networks: a high-resolution network (HRN) and low-resolution network (LRN), with network parameters $W_H$ and $W_L$, respectively. Each network, with parameters W, produced class-membership probabilities ($P(C|x, W)$) for classes $C$ given the gesture?s observation $x$. We multiplied the class-membership probabilities from the two networks element-wise to compute the final class-membership probabilities for the gesture classifier:

$$P(C|x) = P(C|x, W_L) * P(C|x, W_H)$$

We predicted the class label $c* = argmax P(C|x)$. The networks contained more than 1.2 million trainable parameters. The high-resolution network consisted of four 3D convolution layers, each of which was followed by the max-pooling operator. Fig. 1 shows the sizes of the convolution kernels, volumes at each layer, and the pooling operators. We input the output of the fourth 3D convolutional layer to two fully-connected layers (FCLs) with 512 and 256 neurons, respectively. The output of this high-resolution network was a softmax layer, which produced class-membership probabilities ($P(C|x, W_H)$) for the 19 gesture classes. Input the spatially down sampled gesture volume of 28 x 28 x 3 interleaved depth and image gradient values to the low-resolution network. Similar to the HRN and LRN also comprised of a number of 3D convolutional layers, each followed by a max-pooling layer and two FCLs and output softmax layer that estimated the class- membership probability $P(C|x, W_L)$ values. All the layers in the networks, except for the softmax layers, had the rectified linear unit (ReLU) activation functios:

$$f(z) = max(0, z)$$

We computed the output of the softmax layers as:

$$P(C|x, W) = \frac{exp(z_C)}{\sum_q exp(z_q)}$$

where $z_q$ was the output of the neuron $q$.

The process of training a CNN involves the optimization of the network's parameters $W$ to minimize the cost function. The negative log-likelihood is selected as the cost function:

$$L(W, D) = -\frac{1}{|D|} \sum_{i=0}^{|D|} log(P(C^{(i)}|x^{(i)}, W))$$

The optimization is performed via stochastic gradient descent with mini-batches of 40 and 20 training samples for the LRN and HRN respectively. The network's parameters are updated, $W$ with the Nesterov accelerated gradient at every iteration $i$ as:

$$\nabla w_i = \left.\frac{\delta L}{\delta(w_{i-1})}\right|_{batch},$$
$$v_i = \mu v_{i-1} - \lambda \nabla w_i$$
$$w_i = w_{i-1} + \mu v_i - \lambda \nabla w_i$$

where $\lambda$ was the learning rate, $\mu$ was the momentum coefficient, $\nabla w_i$ was the value of gradient of the cost function with respect to the parameter $w_i$ averaged across the mini-batch. The momentum is set to 0.9. The NAG is observed to be converged faster than gradient descent with only momentum.

```
Layer (type)                     Output Shape          Param #     Connected to
====================================================================================================
convolution2d_1 (Convolution2D)  (None, 32, 26, 26)    320         convolution2d_input_1[0][0]
____________________________________________________________________________________________________
convolution2d_2 (Convolution2D)  (None, 64, 24, 24)    18496       convolution2d_1[0][0]
____________________________________________________________________________________________________
maxpooling2d_1 (MaxPooling2D)    (None, 64, 12, 12)    0           convolution2d_2[0][0]
____________________________________________________________________________________________________
dropout_1 (Dropout)              (None, 64, 12, 12)    0           maxpooling2d_1[0][0]
____________________________________________________________________________________________________
flatten_1 (Flatten)              (None, 9216)          0           dropout_1[0][0]
____________________________________________________________________________________________________
dense_1 (Dense)                  (None, 128)           1179776     flatten_1[0][0]
____________________________________________________________________________________________________
dropout_2 (Dropout)              (None, 128)           0           dense_1[0][0]
____________________________________________________________________________________________________
dense_2 (Dense)                  (None, 10)            1290        dropout_2[0][0]
====================================================================================================
Total params: 1,199,882
Trainable params: 1,199,882
Non-trainable params: 0
```

Figure 1: CNN model

The weights of the 3D-convolution layers are initialized with random samples from a uniform distribution between $[-W_b, W_b]$, where $W_b = 6/(n_i + n_o)$, and $n_i$ and $n_o$ were the number of input and output neurons, respectively. We initialized the weights of the fully-connected hidden layers and the softmax layer with random samples from a normal distribution $N(0, 0.01)$. The biases for all layers, except for the softmax layer, were initialized with a value of 1 in order to have a non-zero partial derivative. For the softmax layers biases were set to 0.

### 3.5 Prediction

The general idea of the prediction of the movements is from the generative adversial networks. The Compositional Pattern Producing Networks are a way to represent an entire image as a function. Since neural networks are universal function approximators, given a large enough network, any image of finite resolution can be represented using this method. A simple Variational autoencoder is implemented. Each of the outputs is taken to be the mean of a Bernoulli distribution. The variational lower bound is given by:

$$L = E_{z\ Q(z|x)} log P(X|z) - D(Q(z|X)||P(z))$$

where $P$ is a Bernoulli distribution

The KL divergence between a gaussian $Q$ with mean $\mu$ and standard deviation $\sigma$ and a standard normal distribution $P$ is given by:

$$D(Q||P) = -\frac{1}{2} \sum_{j}^{J} \left[ 1 + log((\sigma_j)^2) - (\mu_j)^2 - (\sigma_j)^2 \right]$$

We want to maximize this lower bound, but because tensorflow doesn't have a 'maximizing' optimizer, we minimize the negative lower bound.

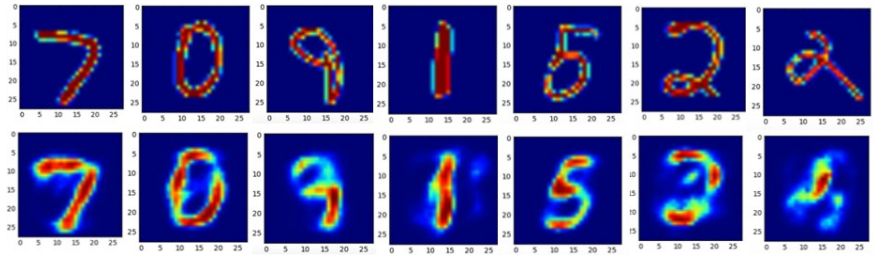

Figure 2: GANs

The most important thing here is to predict the gestures along with the tracking algorithms. Predicting the gestures just with the initial movements

# 4 Results

The image segmentation followed by the detection of the region of interest and tracking are the three important stages in this model. As mentioned earlier, detection is achieved using the segmentation and filters and the tracking by the CamShift algorithm. Finally the classification is implemented using the convolutional neural networks. Figure.5 shows the convex hull detection and the localization of the hand in the frame. It is shown that the localization and the contouring of the area of interest are able to perform well on all lighting conditions.

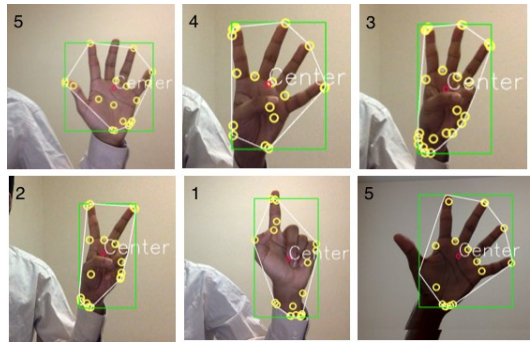

Figure 3: Localization of the hand using the convex-hull algorithm

Next stage is the detection of the gestures and the classification of the gestures. The first step is to convert the real-time images to the binary images after thresholding. Contours are found eventually and area of region shows the difference between the current frame (t) and the previous frame (t-1).

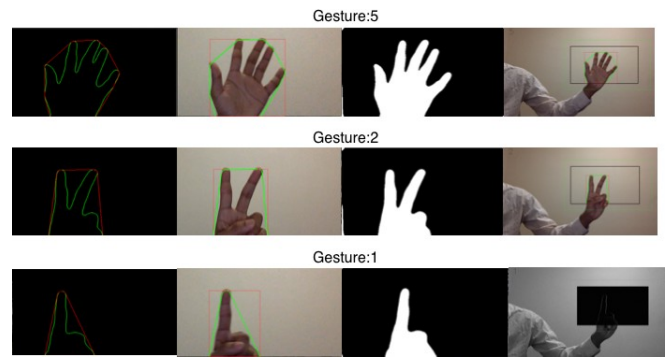

Figure 4: Detection of the gestures: (a) contours of the gestures (b) shows the edges of the fingers (c) binary image of the area of interest (d) shows the image differencing between frame at (t-1) and a frame at (t)

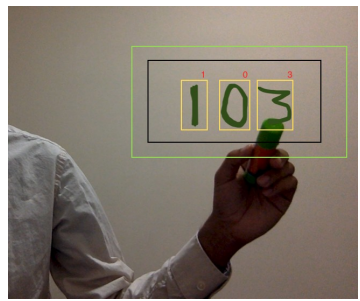

Figure 5: Recognition of the digits

The final figure shows the reconstruction of the images from the variational autoencoder. A VAE is trained with 2d latent space and illustrates how the encoder (the recognition network) encodes some of the labeled inputs (collapsing the Gaussian distribution in latent space to its mean). This gives some insights into the structure of the learned manifold (latent space).

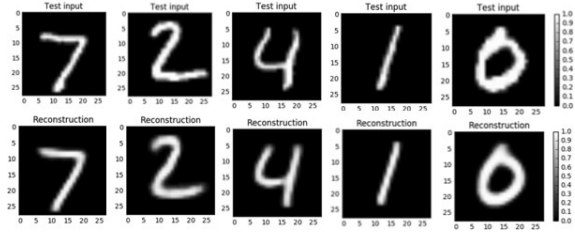

Figure 6: Reconstruction of the images using Generative Adversial Networks VAEs

## 5  Conclusion

The paper presents the model for using the image segmentation, model structures and the number of states on the Gesture Recognition task. It is an effective method for dynamic hand gesture recognition with 3D convolutional neural networks. The proposed classifier uses a fused motion volume of normalized depth and image gradient values, and utilizes spatio-temporal data augmentation to avoid overfitting. By means of extensive evaluation, it is demonstrated that the combination of low and high resolution sub-networks improves classification accuracy considerably. Data augmentation technique plays an important role in achieving superior performance. The future work will include more adaptive selection of the optimal hyperparameters of the CNNs, and investigating robust classifiers that can classify higher level dynamic gestures including activities and motion contexts. The system is fully automatic and it works in real-time. It is fairly robust to background cluster. The advantage of the system lies in the ease of its use. The users do not need to wear a glove, neither is there need for a uniform background. Currently, the tracking method is limited to 2D. Future works include incorporating the prediction method to the image segmentation and the tracking.

## References

[1] Nguyen Dang Binh, & Enokida Shuichi, & Toshiaki Ejima (2005) Real-Time Hand Tracking and Gesture Recognition System

[2] Nianjun Liu, & Brian C. Lovell, & Peter J. Kootsookos, & Richard I.A. Davis, Model Structure Selection & Training Algorithms for a HMM Gesture Recognition System

[3] Pavlo Molchanov, & Shalini Gupta, & Kihwan Kim, & Jan Kautz (2015) Hand Gesture Recognition with 3D Convolutional Neural Networks

[4] D. Ciresan, & U. Meier, & J. Schmidhuber (2012) Multi-column deep neural networks for image classification.